\documentclass[11pt]{article}
\usepackage[left=1in,top=1in,right=1in,bottom=1in]{geometry}
\usepackage{times}

\usepackage{url}
\usepackage{verbatim}
\usepackage{amsmath}
\usepackage{amssymb}
\usepackage{amsthm}
\usepackage{rotating}
\usepackage{algorithm}
\usepackage[noend]{algpseudocode}

\usepackage{tabularx}
\usepackage{subfigure}
\usepackage{multirow}
\usepackage{siunitx}

\newtheorem{fundamental}{Fundamental Rule of Poker Strategy}

\usepackage{silence}
\begin{document}
\title{Most Important Fundamental Rule of Poker Strategy}
\author{Sam Ganzfried$^{1}$, Max Chiswick\\ 
$^1$Ganzfried Research
}

\date{\vspace{-5ex}}

\maketitle
\begin{abstract}
Poker is a large complex game of imperfect information, which has been singled out as a major AI challenge problem. Recently there has been a series of breakthroughs culminating in agents that have successfully defeated the strongest human players in two-player no-limit Texas hold 'em. The strongest agents are based on algorithms for approximating Nash equilibrium strategies, which are stored in massive binary files and unintelligible to humans. A recent line of research has explored approaches for extrapolating knowledge from strong game-theoretic strategies that can be understood by humans. This would be useful when humans are the ultimate decision maker and allow humans to make better decisions from massive algorithmically-generated strategies. Using techniques from machine learning we have uncovered a new simple, fundamental rule of poker strategy that leads to a significant improvement in performance over the best prior rule and can also easily be applied by human players.
\end{abstract}

\section{Introduction}
\label{se:intro}
Poker is an extremely challenging game, both for computer AI programs and for humans, due to the large state spaces and imperfect information. Poker has been very popular with humans for centuries, with the most common variant in recent years being no-limit Texas hold 'em. There are two primary reasons for the popularity of poker in comparison to other games. First is that, while being predominantly a game of skill (which allows professional players to achieve long-term success), there is still a significant degree of luck and variance. In a game like chess, a grandmaster player will beat a recreational club-level player 100\% of the time; however in poker, even an amateur or novice player can still win a single hand or session even against a strong professional player (in fact the Main Event at the World Series of Poker in Las Vegas has been won several times by amateur players). The second reason is that the rules are very simple (allowing easy accessibility to novice players), yet perfect mastery is extremely complex. As a famous professional player Mike Sexton stated, ``The name of the game is No Limit Texas hold 'em, the game that takes a minute to learn but a lifetime to master.'' While anyone can understand the rules and start playing very quickly, even the best experts spend many years (for some an entire lifetime) learning and improving.

Humans learn and improve in a variety of ways. The most obvious are reading books, hiring coaches, poker forums and discussions, and simply playing a lot to practice. Recently several popular software tools have been developed, most notably PioSolver,\footnote{\url{https://piosolver.com/}} where players solve for certain situations, given assumptions for the hands the player and opponent can have. This is based on the new concept of \emph{endgame solving}~\cite{Ganzfried15:Endgame}, where strategies are computed for the latter portion of the game given fixed strategies for the \emph{trunk} (which are input by the user). While it has been pointed out that theoretically this approach is flawed in the worst case in imperfect-information games (such as poker)~\cite{Ganzfried15:Endgame}, nonetheless it has played a pivotal role in producing superhuman agents for two-player no-limit Texas hold 'em~\cite{Brown17:Superhuman,Moravcik17b:DeepStack}. Human players can improve from training with such software tools by studying solutions to many different scenarios, but this requires a lot of time and effort, ability to correctly gauge the trunk strategies, and a significant amount of technical savvy.

Learning to improve as a human is complicated further as there are many different variants of poker commonly played, and concepts learned from one may not transfer over to the other. However, certain theoretical concepts are fundamental to poker and can extend throughout different variants, independent of the specific situation (and in some cases even the opponent). For example, the concept of \emph{pot odds} dictates how good one's \emph{equity} needs to be to make a call worthwhile. For example, suppose the opponent bets \$10 into a pot that initially has \$30. If we call and win then we will profit \$10 + \$30 = \$40, and if we call and lose then we profit -\$10. By contrast, note that if we fold to the bet then we will profit \$0. Suppose we think that we have the best hand with probability $p$: then we are indifferent between calling and folding if $40p - 10 (1-p) = 0 \leftrightarrow p = \frac{1}{5}.$ So calling is more profitable than folding if we think that we win over 20\% of the time, and folding is more profitable if we think that we win less than 20\% of the time. This is a clear simple principle that is applicable to all variants of poker and can be easily applied by human players. (Note however that this concept is not a \emph{rule} that specifies what action the player should take on an absolute level; a \emph{range} of hands must be specified for the opponent in order to apply the concept.) While mastering poker is clearly much more challenging than learning a few general concepts such as pot odds, nonetheless certain fundamental rules grounded in math and game theory can be very important, particularly at earlier phases of study.  

Pot odds does not dictate a full rule for how one should play against opponents with potentially unknown strategy. The most widespread rule for this is based on the concept of the \emph{Minimum Defense Frequency} (MDF). This concept, which is often denoted by $\alpha$~\cite{Ankenman06:Mathematics}, specifies the probability needed to call when facing a bet of a certain size in order to ensure that the opponent cannot profitably bluff us with his weakest hands. (It also simultaneously specifies the optimal ratio that a player should use of \emph{value bets} (bets with a strong hand) to \emph{bluffs} (bets with a weak hand) that a player should use.) The minimum defense frequency is derived from the Nash equilibrium of a simplified game, and has been singled out as one of the most important concepts of poker. When facing a bet by a player with an unknown strategy, many strong players will base their strategy off of MDF. This is a simple effective rule that is easily understood by humans and applies to many different variations of poker. In this paper we will develop a new rule that is also simple and can easily be understood by humans that improves performance significantly beyond MDF. This rule integrates MDF with another popular and easily understandable poker concept called \emph{Range Advantage}, and is derived by applying machine learning algorithms to a database of 
solutions to simplified games. We believe that this is the most important single rule for strong poker play. 

Poker is large game of imperfect-information and has been singled out as being a major AI challenge problem. Recently programs for two-player no-limit Texas hold 'em have competed against the strongest human players in a sequence of high-profile ``Brains vs. Artificial Intelligence'' competitions~\cite{Brown17:Superhuman} culminating in the agent Libratus that won decisively in 2017. Independently a second agent DeepStack was also created that defeated human players in 2017, though the humans were not experts in the specific game format~\cite{Moravcik17b:DeepStack}. Despite these successes, there have been relatively few takeaways from the research that aspiring human players can readily apply to improve their poker game. 
In a separate new line of research, a new approach has been developed using decision trees to compute strategic rules that are easily understood by humans~\cite{Ganzfried17b:Computing}. This has led to deduction of new rules concerning when a player should make an extremely small bet and an extremely large bet. We continue this avenue of research by exploring for the first time the connection between two key poker strategy concepts, producing a single understandable rule that applies to a variety of bet sizes and different game variations and significantly improves performance over the popular MDF concept. Large-scale computation of strong game-theoretic strategies is also important in many domains other than poker; for example, it has been applied recently to randomized security check systems for airports~\cite{Paruchuri08:Playing}. Approaches for computing human-understandable rules are critical for situations such as national security where humans will be making important real-time decisions.

\section{No-limit poker}
\label{se:poker}
We will be studying the following simplified poker game. First player 1 is dealt a card from a $n$-card deck (cards are the numbers 1--$n$), according to a probability distribution $p$ (e.g., $p_2$ is the probability that player 1 is dealt a 2). We assume similarly that player 2 is dealt a card from 1--$n$ according to $q$. We assume the cards are independent, and that it is possible for both players to be dealt the same card (so that our experiments will not be confounded by the additional issue of \emph{card removal}). Players initially start with a stack $S$ and there is a pot of size $P$, with a deck size equal to $n = 10$. Then player 1 can either \emph{check} or \emph{bet} a size in $\{b_{1i}\}$. Facing a bet from player 1, player 2 can then call or \emph{fold} (forfeit the pot). Facing a check, player 2 can check or bet a size in $\{b_{2i}\}$. Then player 1 can call or fold facing a bet. If neither player folds, the player with higher card wins the pot.

For our initial experiments we will assume the pot and stack sizes $P$ and $S$ both equal 1 (if they differed we could divide by the size of the pot to normalize).  For both players we only allow a bet size equal to pot (1) for $b_{1i},b_{2i}.$ We will also consider settings where the bet size is equal to 0.5 times the pot and 0.75 times the pot (which are popular human betting sizes). Of course this setting can allow arbitrary values for these parameters and multiple bet sizes, but for concreteness we decided to start with these values.

\begin{figure}[!ht]
\centering
\begin{minipage}{0.32\textwidth}
\centering
\includegraphics[width=\linewidth]{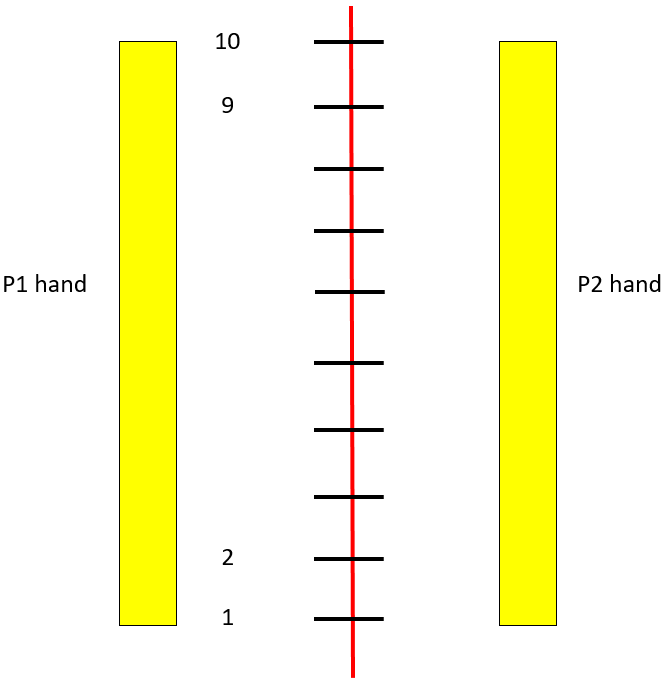}
\caption{Both players are dealt hands uniformly at random over all hands.}
\label{fi:uniform}
\end{minipage}
\hfill
\begin{minipage}{0.32\textwidth}
\centering
\includegraphics[width=\linewidth]{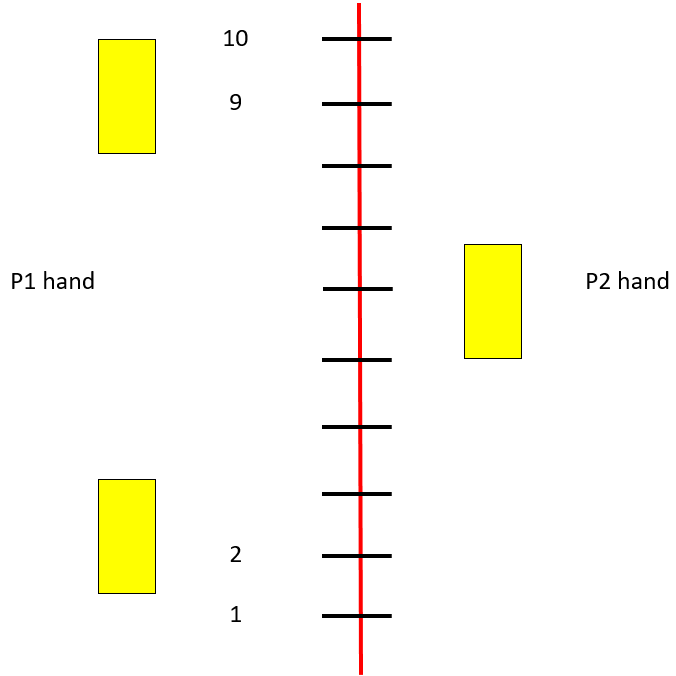}
\caption{Player 1 is dealt either a very strong or weak hand and player 2 is always dealt a mediocre hand.}
\label{fi:p1polar}
\end{minipage}
\end{figure}

It is clear that the optimal strategies (according to Nash equilibrium) for the two players will vary significantly depending on the distributions $p$ and $q$. 
For example, if both players are dealt cards uniformly (Figure~\ref{fi:uniform}), then a Nash equilibrium strategy for player 1's first action is:
\begin{itemize}
\item Card 1: Check pr. 1
\item Card 2: Check pr. 0.5, bet 1 pr. 0.5
\item Card 3-8: Check pr. 1 
\item Card 9: Bet 1 pr. 1 
\item Card 10: Check pr. 1 
\end{itemize}
This can be computed quickly using known techniques. 
Note that the equilibrium strategy includes \emph{bluffing} (i.e., betting with weak hands such as 2), as well as \emph{slowplaying}, aka \emph{trapping} (checking with a strong hand such as 10).  

However, suppose the cards are dealt according to a different distribution: player 1 is either dealt a very strong hand (10) or a very weak hand (1) with probability 0.5 while player 2 is always dealt a medium-strength hand (Figure~\ref{fi:p1polar}). Then the equilibrium strategy for player 1 is:
\begin{itemize}
\item Card 1: Bet 0 pr. 0.5, 1 pr. 0.5 
\item Card 10: Bet 1 pr. 1
\end{itemize}
If player 1 is always dealt a medium-strength hand (5) while player 2 is dealt a very strong or very weak hand with probability 0.5 
then the equilibrium strategy is:
\begin{itemize}
\item Card 5: Bet 0 pr. 1 
\end{itemize}

\section{Qualitative models and endgame solving}
\label{se:qual}
\begin{figure}[!ht]
\centering
\begin{minipage}{0.32\textwidth}
\centering
\includegraphics[width=\linewidth]{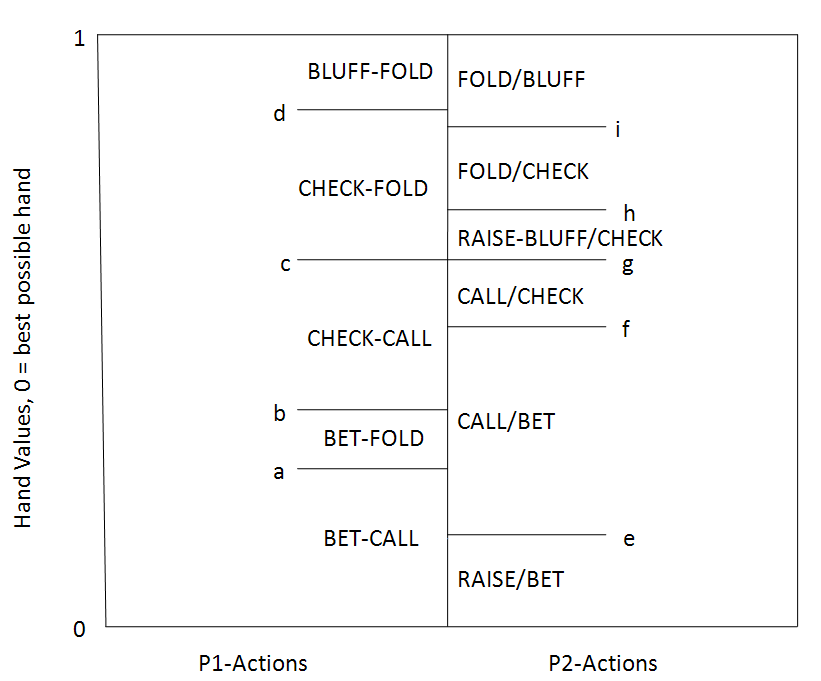}
\end{minipage}
\hfill
\begin{minipage}{0.32\textwidth}
\centering
\includegraphics[width=\linewidth]{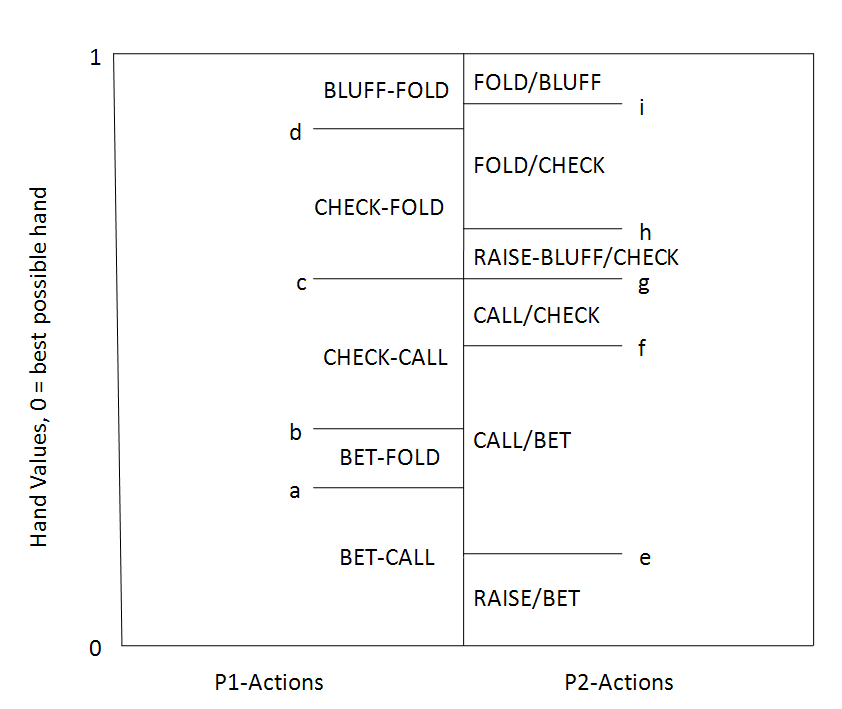}
\end{minipage}
\hfill
\begin{minipage}{0.32\textwidth}
\centering
\includegraphics[width=\linewidth]{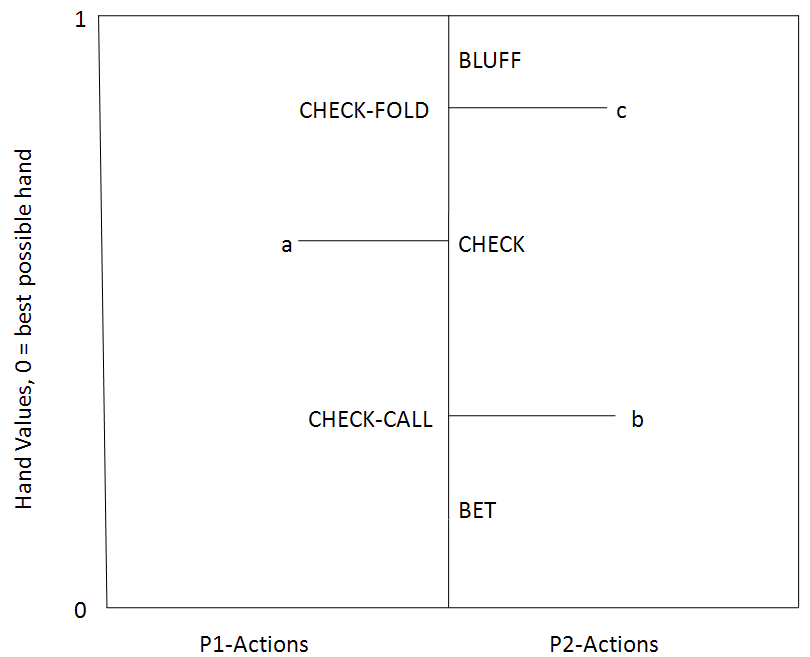}
\end{minipage}
\caption{Three qualitative models for two-player limit Texas hold 'em river endgame play.}
\label{fi:qualitative-models}
\end{figure}

There has been some prior study of human understandable strategies in imperfect-information games, and in poker specifically. Ankenman and Chen compute analytical solutions of several simplified poker variants by first assuming a given qualitative structure on the equilibrium strategies, and then computing strategies given this presumed structure~\cite{Ankenman06:Mathematics}. While the computed strategies are generally interpretable by humans, the models were typically constructed from a combination of trial and error and expert intuition, not algorithmically. More recent work has shown that leveraging such qualitative models can lead to new equilibrium-finding algorithms that outperform existing approaches~\cite{Ganzfried10:Computing}. That work proposed three different qualitative models for the final round of two-player limit Texas hold 'em (Figure~\ref{fi:qualitative-models}), and showed empirically that equilibrium strategies conformed to one of the models for all input information distributions (and that all three were needed). Again here the models were constructed by manual trial and error, not algorithmically.

We note that while the problem we are considering in this paper is a ``toy game,'' it captures important aspects of real poker games and we expect our approaches to have application to larger more realistic variants (our game generalizes many common testbed variants; 
if the deck had only 3 cards and only one bet size is allowed, this would be the commonly-studied variant ``Kuhn poker''~\cite{Kuhn50:Simplified}). In the recent Brains vs. Artificial Intelligence two-player no-limit Texas hold 'em competition, the agent Claudico computed the strategy for the final betting round in real time, and 
a top human player has commented that the ``endgame solver'' was the strongest component of the agent~\cite{Ganzfried17:Reflections}; endgame solving was also a crucial component of subsequent success of the improved agent Libratus~\cite{Brown17:Superhuman}. 
Another recent superhuman agent DeepStack can also be viewed as applying endgame solving, as it works by viewing different rounds of the game as separate endgames which are solved independently, using deep learning to estimate the values of the endgames terminal states~\cite{Moravcik17b:DeepStack}. Endgame solving assumes that both agents had private information distributions induced by the strategies for the prior rounds using Bayes' rule, assuming they had been following the agent's strategy for the prior rounds~\cite{Ganzfried15:Endgame}. The game we study here is very similar to no-limit Texas hold 'em endgames, except that we are assuming a ten-card deck, specific stack and betting sizes, and that raises are not allowed. We expect our analysis to extend in all of these dimensions and that our approaches will have implications for no-limit Texas hold 'em and many other complex variants. 

\section{\large Minimum defense frequency and range advantage}
\label{se:mdf}
For a bet of size $b$ and a pot of size $P$, the \emph{minimum defense frequency} (MDF) is defined to be the quantity $\alpha = \frac{P}{b+P}.$ Suppose player 2 calls the bet with probability $\alpha$, and suppose that player 1 has an extremely weak hand that he is contemplating betting as a bluff. With probability $\alpha$ the bluff would be called and player 1 would lose $b$, while with probability $1 - \alpha$ player 2 would fold and player 1 would win $P$. Thus the expected profit for player 1 from betting is: 
\small
$$\alpha \cdot (-b) + (1-\alpha)P = P - \alpha (b+P) = P -  \frac{P}{b+P} \cdot (P+b) = 0.$$
\normalsize
Thus player 1 is precisely indifferent between betting as a bluff and not betting with his weak hand. If player 2 were to call with any probability below $\alpha$, then player 1 would be able to have a guaranteed profit by bluffing with all his weak hands; therefore, the value $\alpha$ is the minimum amount of the time that player 2 must call (or raise if it is permitted) against a bet in order to prevent player 1 from profitably betting with all his weak hands. For our parameters of $b = 1, P=1$ the MDF is $\alpha = \frac{1}{2}.$ Note also that if player 2 calls with probability exceeding $\alpha$, then player 1 would never want to bluff with his weak hands (in which case player 2 would never want to call with his medium-strength hands, etc.). So in general we would expect player 2 to call with frequency precisely equal to $\alpha$; however this is not always the case. Consider an extreme example when player 1 is always dealt a 10 and player 2 is always dealt a 1. Then clearly player 2 would always prefer to fold to a bet because he will never have the best hand. Note that in this example player 1 has a \emph{range advantage} equal to 1. In general, if player 1 has a significant range advantage, optimal play 
may suggest that player 2 call with probability below $\alpha$. 

The \emph{range advantage} for player 1 under the distributions $p,q$ is player 1's \emph{equity} in the hand given the distributions. This is equal to the probability that player 1 will have a better hand under $p$ plus one-half times the probability that they have the same hand. So we define the Range Advantage as: 

\small
\begin{equation} 
\mbox{Range Advantage}(p,q) = \left(\sum_{j = 1}^n \sum_{i = j+1}^n p_i q_j\right) + \left(\frac{1}{2} \sum_{i=1}^n p_i q_i\right)
\end{equation}
\normalsize

We generated 100,000 games with uniform random distributions for $p$ and $q$ (recall that $p$ is player 1's distribution and $q$ is player 2's). We then solve these games for a Nash equilibrium. Let $c(i)$ be the optimal frequency for player 2 to call a bet from player 1 with card $i$ according to the equilibrium strategy. Then the \emph{optimal defense frequency} (ODF) $c^*$ is the weighted sum of the call probabilities over all hands, i.e., $c^* = \sum_i \left (q_i c(i) \right).$ In our experiments we seek to understand the relationship between the range advantage and the optimal defense frequency which, as described above, may differ considerably from the minimum defense frequency.     

\begin{figure}[!ht]
\centering
\includegraphics[width=0.9\linewidth]{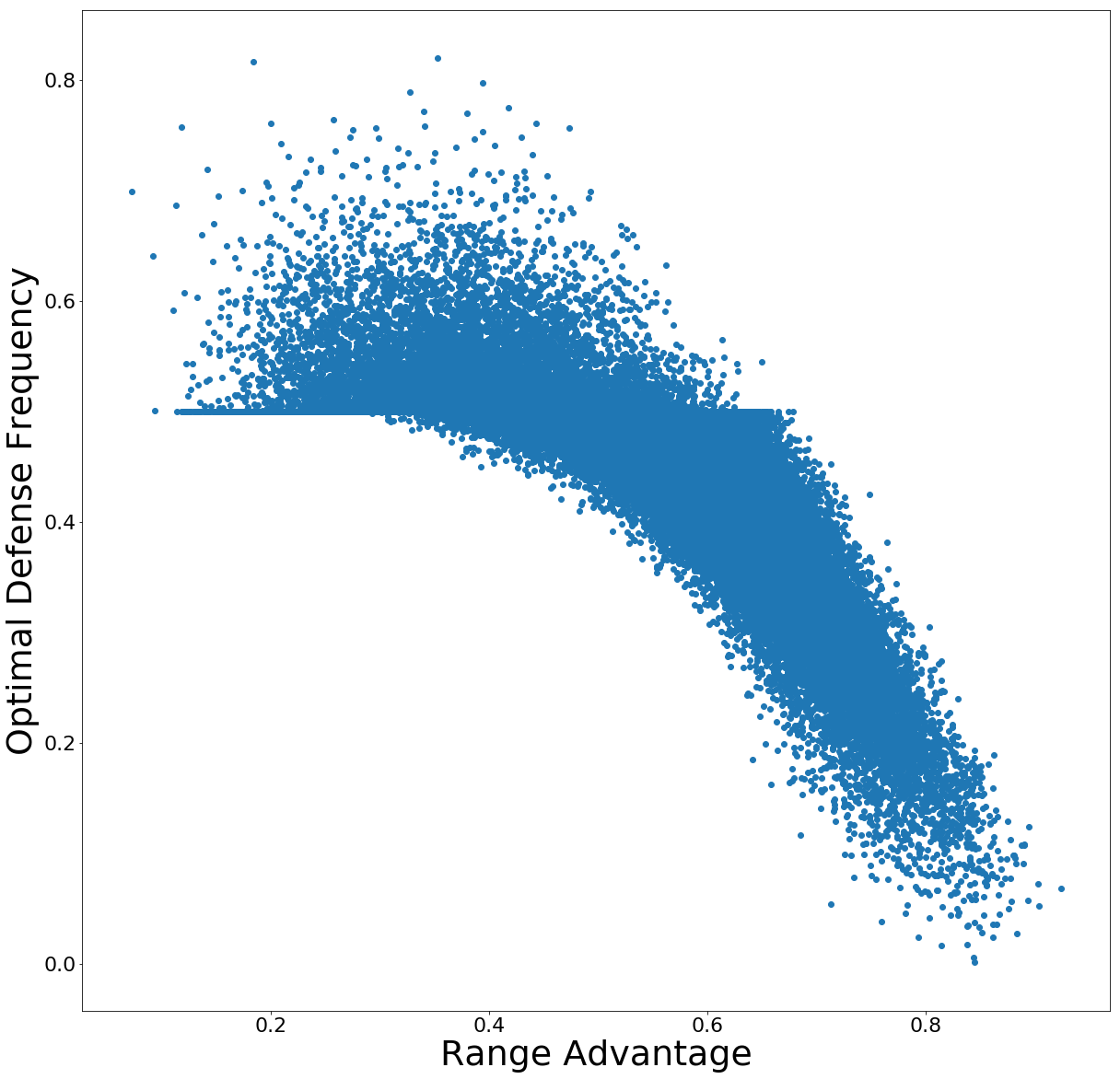}
\caption{Range advantage vs. optimal defense frequency over 100,000 random games with bet size equal to pot.}
\label{fi:scatter}
\end{figure}

For each hand in the sample, we can add a data point where the x coordinate is the range advantage of the distributions $p,q$, and the y coordinate is the optimal defense frequency $c^*$. A scatterplot of points selected according to this process is shown in Figure~\ref{fi:scatter}. The graph depicts a clear negative trend between the range advantage of player 1 and the optimal defense frequency for player 2
. We also observe a mass of hands with optimal frequency exceeding the MDF value of 0.5.  We note that each game may contain several solutions, and it is possible that there also exist other equilibria with different ODF values. Thus, for the games with ODF exceeding 0.5, it is possible that the game also contains an equilibrium with ODF equal to 0.5, and that our algorithm happened to select this one. This situation can arise for games where player 1's equilibrium strategy checks with all hands and never bets; in this case there are many equilibrium strategies for player 2 to play in response to a hypothetical ``off-path'' bet action by player 1, and player 2 need only play a strategy that does not give player 1 incentive to deviate and start betting some hands instead of checking.

\section{Experiments}
We first compare several approaches for the variant where the players are restricted to just the pot-sized bet: ``previous expert knowledge [has] dictated that if only a single bet size [in addition to all-in] is used everywhere, it should be pot sized''~\cite{Hawkin12:Using}.

For each approach, we compute the mean-squared error (MSE) from the optimal defense frequency (ODF) over the 100,000 game sample 
First we consider just predicting the MDF, which is 0.5 for this case of a pot-sized bet, which produces MSE of 0.0067. We next considered ``Linear MDF,'' which in this case would select the optimal constant predictor, which selected 0.466 as opposed to 0.5, and produced MSE of 0.0055. We next integrated range advantage as a feature to produce the optimal regressor prediction of -.446*RA + 0.688, with MSE 0.0024, and finally quadratic regression which produced MSE of 0.0012. These results (Table~\ref{ta:exp-pot}) show that adding in range advantage as a feature leads to a significant reduction in MSE over just using MDF (around a 56\% reduction even accounting for optimized normalization of MDF), while also unsurprisingly quadratic leads to a further reduction. Though there is a risk of overfitting, the plot in Figure~\ref{fi:scatter} appears to fit closer with a quadratic than a linear curve, so the quadratic predictor seems more appropriate. For all approaches we perform 10-fold cross validation.

\renewcommand{\tabcolsep}{2pt}
\begin{table}[!ht]
\centering
\small
\begin{tabular}{|*{3}{c|}} \hline
\textbf{Approach} &\textbf{Formula} &\textbf{MSE} \\ \hline
Fixed MDF &0.5 &0.006686 \\ \hline
Linear MDF &0.466 &0.005512 \\ \hline
Linear RA &-.446*RA + 0.688 &0.002406 \\ \hline
Quadratic RA &-1.760*RA$^2$ + 1.316*RA + 0.275 &0.001161 \\ \hline
\end{tabular}
\caption{Results for simplified game with pot bet size.}
\label{ta:exp-pot}
\end{table}

We next ran experiments over a larger dataset consisting of 100,000 games for each of the three bet sizes of 0.5 pot, 0.75 pot, and pot. 
These experiments allowed us to uncover a rule that would extend beyond different bet sizes, as the different bet sizes correspond to different values of the MDF $\alpha$. For these experiments the MDF now becomes a feature (while it was just the constant 0.5 for the prior experiments). The MSE of several approaches are given in Table~\ref{ta:exp-multi}. The MSE of fixed MDF is 0.0086, which is reduced to 0.0068 for linear MDF (note that, in contrast to the pot-sized experiments where MDF was a fixed constant, in this setting where MDF is a feature there is now both a coefficient for MDF and a constant parameter). Using both MDF and RA brought down MSE further to 0.0030 (similar to the pot-sized experiments including RA as a feature reduces MSE over linear MDF by around 56\%). We also used the product of RA and MDF as a new feature, and including this did not lead to a significant reduction in MSE over just using RA (both when used in addition to RA and in place of it).

\renewcommand{\tabcolsep}{2pt}
\begin{table*}[!ht]
\centering
\scriptsize
\begin{tabular}{|*{3}{c|}} \hline
\textbf{Approach} &\textbf{Formula} &\textbf{MSE} \\ \hline
Fixed MDF &MDF &0.008586 \\ \hline
Linear MDF &0.904*MDF + 0.0140 &0.006791 \\ \hline
Linear MDF with RA &0.904*MDF - 0.495*RA + 0.261 &0.002996\\ \hline
Linear MDF with RA*MDF &1.330*MDF - 0.851*[RA*MDF] + 0.0134 &0.002971 \\ \hline
Linear MDF with RA and RA*MDF &1.202*MDF - 0.150*RA - 0.595*[RA*MDF] + 0.0885 &0.002966 \\ \hline
Simplified Linear MDF with RA &MDF - 0.5*RA + 0.25 &0.004795 \\ \hline
Min Linear MDF with RA &min(MDF, 0.904*MDF - 0.495*RA + 0.261) &0.002638 \\ \hline
\textbf{Simplified Min Linear MDF with RA} &\textbf{min(MDF, MDF - 0.5*RA + 0.25)} &\textbf{0.003196} \\ \hline
Piecewise Simp. Min Linear MDF w. RA &1 - RA, RA $>$ 0.75; min(MDF, MDF - 0.5*RA + 0.25), RA $\leq$ 0.75 &0.002446 \\ \hline
Quadratic MDF with RA and MDF*RA &-1.945*RA$^2$ - 0.615*RA*MDF + 0.0111*MDF$^2$ + 1.806*RA + 1.197*MDF - 0.368 &0.001459 \\ \hline
\end{tabular}
\caption{Results for several approaches over 100,000 hands for each of three games with bet sizes of 0.5 pot, 0.75 pot, and pot.}
\label{ta:exp-multi}
\end{table*}

The optimal linear predictor using RA and MDF (rounded to 3 decimal places) is 0.904 * MDF - 0.495 * RA + 0.261. Noting that the first coefficient is close to 1, the second is close to 0.5, and the third is close to 0.25, we also investigated this ``simplified'' predictor with coefficients that can be more easily remembered and applied in real time. This produced an MSE of 0.0048, which is about halfway between the MSE of Linear MDF and Linear MDF with RA. 

Another observation we made from Figure~\ref{fi:scatter} is that the optimal defense frequency only rarely exceeds the MDF (and furthermore the pattern for the points where ODF exceeds MDF seems to be different from that of the points for which it doesn't). So we decided to consider an approach where we truncate the predictor at the MDF (i.e., take the minimum of the predictor value and MDF). We also considered an approach where we used a separate simple linear function for the cases where player 1 has a significant range advantage (RA $>$ 0.75) of 1-RA for RA values exceeding 0.75, while using the same approach as before for RA $\leq$ 0.75.

Using min(MDF, 0.0904*MDF - 0.495*RA + 0.261) produced MSE of 0.0026, which is only a small improvement over that of linear MDF with RA (without using the min function). However, applying the min function to the simplified linear predictor of MDF - 0.5*RA + 0.25 led to a significant reduction in MSE from 0.0048 to 0.0032 (33\% reduction). Adding in the separate rule of using (1-RA) for RA $>$ 0.75 further decreased MSE to 0.0024. Finally we note that quadratic regression not surprisingly produced the lowest MSE out of the rules we considered. There are many more complex approaches we could have also experimented with (e.g., additional features, decision trees, higher degree regression, etc.), however these run a significant risk of overfitting (and also producing less-understandable rules).

\section{New fundamental rule of poker strategy}
\label{se:fundamental}
Many of the approaches presented in Table~\ref{ta:exp-multi} achieve a significant performance improvement over just (linear) MDF. While not the best-performing approach, we single out the rule min(MDF, MDF - 0.5*RA + 0.25) as being the best-performing single rule using simple coefficients that are easy for a human to remember and apply. This rule produces MSE of 0.0032, which is over a 50\% reduction in MSE over linear MDF (and a 63\% reduction in MSE over fixed MDF). This rule is a single equation, with very simple values of the coefficients (1, -0.5, and 0.25). It only performs slightly worse than the analogous approaches that use more complicated coefficient values. It is also outperformed by a piecewise approach that adds in a second case of (1-RA) for the case RA $>$ 0.75, and uses the same predictor for RA $\leq 0.75$. However, we consider this approach as essentially two different rules combined into one approach, not as a single rule.

\begin{fundamental}
Given minimum defense frequency value MDF when facing a certain bet size, and assuming a range advantage of RA, then you should call the bet with a fraction of the hands in your range equal to min(MDF, MDF - 0.5*RA + 0.25), and fold otherwise.
\end{fundamental}

Note that for RA = 0 (player 2 always has the best hand), this will predict ODF = min(MDF, MDF + 0.25) = MDF. For RA = 1 it will predict ODF = min(MDF, MDF -0.25) = MDF - 0.25. And for RA = 0.5 (neither player has a range advantage) ODF = min(MDF, MDF) = MDF (which is expected for the case when no player has a clear advantage).  

As an example to apply this rule, suppose we are in a setting where opponent bets the pot (so MDF = 0.5), and we believe that the opponent has a range advantage of 0.8. Then we have min(MDF, MDF - 0.5*RA + 0.25) = min (MDF, MDF - 0.15) = MDF - 0.15 = 0.5-0.15 = 0.35.
So the rule would stipulate that we should call 35\% of the time. 

Recall that previously the best rule to apply here would be just to use ODF = MDF = 0.5. Our new rule produces an improvement of over 60\% over the best prior approach, while being a simple rule that can be easily understood and applied in real time. Had we instead defined the range advantage to range from -1 to 1 instead of 0 to 1 (so that RA of 0 meant no advantage and +1 means player 1 has full advantage), then the rule becomes min(MDF, MDF - 0.25*RA), which is even easier for a human to interpret. This is obtained by applying the transformation $RA_{\mbox{new}} \leftarrow 2*RA_{\mbox{old}} - 1$. 

We can call this rule the ``100--50--25 MIN rule,'' denoting the fact that the three coefficients (1, -0.5, 0.25) are in the proportions of 100--50--25 and application of the MIN function. As this rule obtains significantly lower MSE than the prior best rule of using MDF, the ``100--50--25 MIN rule'' is the most important fundamental rule of poker strategy.

We note that several prior ``fundamental rules'' have been proposed in the literature, which are based primarily on anecdotes and intuition. For example in one popular book on strategy, Phil Gordon writes, ``\textbf{Limping is for Losers.} This is \emph{the most important fundamental} in poker---for every game, for every tournament, every stake: If you are the first player to voluntarily commit chips to the pot, open for a raise. Limping is inevitably a losing play. If you see a person at the table limping, you can be fairly sure he is a bad player.
''~\cite{Gordon11:Phil} (Note that some very strong players, e.g., poker AI Claudico (which is Latin for ``I limp''), do in fact ``limp'' sometimes).\footnote{A \emph{limp} is a conservative play of just calling the minimum bet in the first round as opposed to the more aggressive raise action.}
~\cite{Ganzfried17:Reflections} Another rule called ``Zeebo's Theorem'' states: ``No player is capable of folding a full house on any betting round, regardless of the size of the bet.''\footnote{\url{http://www.thepokerbank.com/strategy/theorems/zeebo/}} 
We think our rule is more important than these and furthermore is based on rigorous analysis.


\section{Conclusion}
\label{se:conc}
Using techniques from machine learning, we have been able to obtain a new important rule of poker strategy that achieves significantly lower MSE than the most popular rule of basing calling decisions on MDF. We obtain the 100--50--25 MIN rule, which is a single easily-understandable equation for the predictor of ODF, which produces an improvement of over 60\% of the best prior approach. Generating improved rules or features can enhance performance of machine learning algorithms. Of course even better rules could be obtained (we obtained some, for example with more complex coefficients, piecewise linear functions that branch based on RA value, and quadratic regression); however these are all more complex and would be harder for a human player to memorize and implement in real time. There can also be some concern about overfitting with more complex rules.
It would be interesting to explore implications for the recent line of work on developing approaches for computing human-understandable strategies to other domains where game-theoretic algorithms produce strategies that must be interpreted by human decision makers. For example, in national security humans often ultimately make the decisions that may be generated by algorithms~\cite{Paruchuri08:Playing}. 


\bibliographystyle{plain}
\bibliography{C://FromBackup/Research/refs/dairefs}

\begin{thebibliography}{10}

\bibitem{Ankenman06:Mathematics}
Jerrod Ankenman and Bill Chen.
\newblock {\em The Mathematics of Poker}.
\newblock {ConJelCo LLC}, Pittsburgh, PA, USA, 2006.

\bibitem{Brown17:Superhuman}
Noam Brown and Tuomas Sandholm.
\newblock Superhuman {AI} for heads-up no-limit poker: Libratus beats top
  professionals.
\newblock {\em Science}, 359:418--424, 2017.

\bibitem{Ganzfried17:Reflections}
Sam Ganzfried.
\newblock Reflections on the first man vs. machine no-limit {T}exas hold 'em
  competition.
\newblock {\em AI Magazine}, 38(2), 2017.

\bibitem{Ganzfried10:Computing}
Sam Ganzfried and Tuomas Sandholm.
\newblock Computing equilibria by incorporating qualitative models.
\newblock In {\em Proceedings of the International Conference on Autonomous
  Agents and Multi-Agent Systems (AAMAS)}, 2010.

\bibitem{Ganzfried15:Endgame}
Sam Ganzfried and Tuomas Sandholm.
\newblock Endgame solving in large imperfect-information games.
\newblock In {\em Proceedings of the International Conference on Autonomous
  Agents and Multi-Agent Systems (AAMAS)}, 2015.

\bibitem{Ganzfried17b:Computing}
Sam Ganzfried and Farzana Yusuf.
\newblock Computing human-understandable strategies: {D}educing fundamental
  rules of poker strategy.
\newblock {\em Games}, 8(4):1--13, 2017.

\bibitem{Gordon11:Phil}
Phil Gordon.
\newblock {\em Phil Gordon's Little Gold Book: Advanced Lessons for Mastering
  Poker 2.0}.
\newblock Gallery Books, 2011.

\bibitem{Hawkin12:Using}
John Hawkin, Robert Holte, and Duane Szafron.
\newblock Using sliding windows to generate action abstractions in
  extensive-form games.
\newblock In {\em Proceedings of the AAAI Conference on Artificial Intelligence
  (AAAI)}, 2012.

\bibitem{Kuhn50:Simplified}
H.~W. Kuhn.
\newblock Simplified two-person poker.
\newblock In H.~W. Kuhn and A.~W. Tucker, editors, {\em Contributions to the
  Theory of Games}, volume~1 of {\em Annals of Mathematics Studies, 24}, pages
  97--103. Princeton University Press, Princeton, New Jersey, 1950.

\bibitem{Moravcik17b:DeepStack}
Matej Morav{\v c}{\'\i}k, Martin Schmid, Neil Burch, Viliam Lis{\'y}, Dustin
  Morrill, Nolan Bard, Trevor Davis, Kevin Waugh, Michael Johanson, and Michael
  Bowling.
\newblock Deepstack: Expert-level artificial intelligence in heads-up no-limit
  poker.
\newblock {\em Science}, 356:508--513, 2017.

\bibitem{Paruchuri08:Playing}
Praveen Paruchuri, Jonathan~P. Pearce, Janusz Marecki, Milind Tambe, Fernando
  Ordonez, and Sarit Kraus.
\newblock Playing games with security: An efficient exact algorithm for
  {B}ayesian {S}tackelberg games.
\newblock In {\em Proceedings of the International Conference on Autonomous
  Agents and Multi-Agent Systems (AAMAS)}, 2008.

\end{thebibliography}

\end{document}